\documentclass[11pt, a4paper]{article}
\usepackage{fullpage}
\usepackage{amsfonts}
\usepackage{amssymb}
\usepackage{amsmath}
\usepackage{amsthm}
\usepackage{graphicx}
\usepackage{bm}
\usepackage[makeroom]{cancel}
\usepackage{enumitem}
\usepackage{url}
\usepackage[margin=.9in]{geometry}
\usepackage{amsopn}
\usepackage{mathtools}
\usepackage{hyperref}
\usepackage{doi}
\usepackage{cite}
\usepackage{overpic}
\usepackage[symbol]{footmisc}

\newcommand{\by}{\mathbf{s}}
\newcommand{\bW}{\mathbf{W}}
\newcommand{\bx}{\mathbf{x}}

\DeclareMathOperator*{\argmin}{arg\,min}

\usepackage{listings}
\usepackage{xcolor}
\definecolor{codegreen}{rgb}{0,0.6,0}
\definecolor{codegray}{rgb}{0.5,0.5,0.5}
\definecolor{codepurple}{rgb}{0.58,0,0.82}
\definecolor{backcolour}{rgb}{0.95,0.95,0.92}
\lstdefinestyle{mystyle}{
    backgroundcolor=\color{backcolour},   
    commentstyle=\color{codegreen},
    keywordstyle=\color{magenta},
    numberstyle=\tiny\color{codegray},
    stringstyle=\color{codepurple},
    basicstyle=\ttfamily\normalsize,
    breakatwhitespace=false,         
    breaklines=true,                 
    captionpos=b,                    
    keepspaces=true,                 
    numbers=left,                    
    numbersep=3pt,                  
    showspaces=false,                
    showstringspaces=false,
    showtabs=false,                  
    tabsize=4,
}
\begin{document}
\begin{center}
    \Large \bf PySHRED: A Python package for SHallow REcurrent Decoding for sparse sensing, model reduction and scientific discovery
\end{center}
\begin{center}
    David Ye$^1$\footnote[1]{Corresponding authors (pyshred1@gmail.com)},
    Jan Williams$^{3}$,
    Mars Gao$^{2}$,
   Stefano Riva$^{4}$, 
    Matteo Tomasetto$^{5}$,\\
    David Zoro$^{6}$,
    J. Nathan Kutz$^{6,7}$
\end{center}
\begin{center}
    \scriptsize{
    ${}^1$ Department of Applied Computational Mathematical Sciences, University of Washington, Seattle, WA 98195, United States \\  
    ${}^2$ Department of Computer Science, University of Washington, Seattle, WA 98195, United States \\ 
    ${}^3$ Department of Mechanical Engineering, University of Washington, Seattle, WA 98195, United States \\ 
   ${}^4$ Department of Energy, Nuclear Engineering Division, Politecnico di Milano, Milan, Italy \\
   ${}^5$ Department of Mechanical Engineering, Politecnico di Milano, Milan, Italy
   \\
    ${}^6$ Department of Electrical and Computer Engineering, University of Washington, Seattle, WA 98195, United States \\
     ${}^7$ Department of Applied Mathematics, University of Washington, Seattle, WA 98195, United States     }
\end{center}

\begin{abstract}
\textbf{SH}allow \textbf{RE}current \textbf{D}ecoders (SHRED) provide a deep learning strategy for modeling high-dimensional dynamical systems and/or spatiotemporal data from dynamical system snapshot observations. PySHRED is a Python package that implements SHRED and several of its major extensions, including for robust sensing, reduced order modeling and physics discovery. In this paper, we introduce the version 1.0 release of PySHRED, which includes data preprocessors and a number of cutting-edge SHRED methods specifically designed to handle real-world data that may be noisy, multi-scale, parameterized, prohibitively high-dimensional, and strongly nonlinear. The package is easy to install, thoroughly-documented, supplemented with extensive code examples, and modularly-structured to support future additions. The entire codebase is released under the MIT license and is available at \href{https://github.com/pyshred-dev/pyshred}{https://github.com/pyshred-dev/pyshred}.
\end{abstract}

\section{Introduction}
\begin{figure}[t]
   \centering
   \includegraphics[scale=0.45]{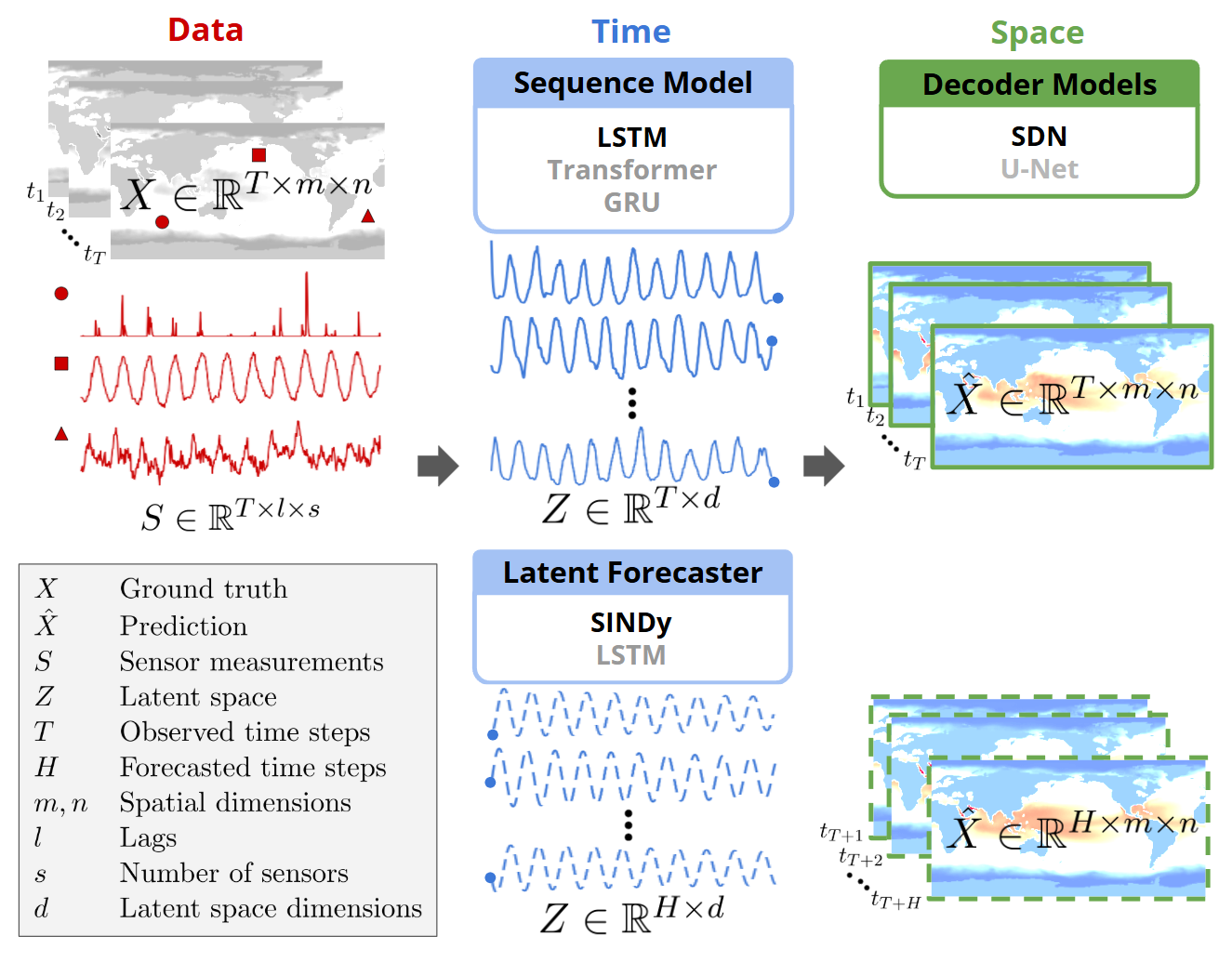}
   \caption{Overview of the SHRED architecture.  SHRED provides a training framework mapping limited temporal (sequence) measurements to high-dimensional dynamical systems and/or spatiotemporal data from dynamical system snapshot observations.}
   \label{fig:pyshred-architecture}
\end{figure}

PySHRED is a Python package that implements the \textbf{SH}allow \textbf{RE}current \textbf{D}ecoder (SHRED) architecture (Figure~\ref{fig:pyshred-architecture}) and provides a high-level interface for sensing, model reduction and physics discovery tasks. Originally proposed as a sensing strategy which is agnostic to sensor placement~\cite{williams2024sensing}, SHRED provides a lightweight, data-driven framework for reconstructing and forecasting high-dimensional spatiotemporal states from sparse sensor measurements. SHRED achieves this by (i) encoding time-lagged sensor sequences into a low-dimensional latent space using a sequence model, and (ii) decoding these latent representations back into the full spatial field via a decoder model.

Since its introduction as a sparse sensing algorithm, several specialized variants have been developed to extend SHRED's capabilities:
\begin{itemize}
  \item \emph{SHRED-ROM} for parametric reduced-order modeling
  \item \emph{SINDy-SHRED} for discovering sparse latent dynamics and stable long-horizon forecasting
  \item \emph{Multi-field SHRED} for modeling dynamically coupled fields
\end{itemize}
PySHRED unifies these variants into a single open-source, extensible, and thoroughly documented Python package, which is also capable of training on compressed representations of the data, allowing for efficient laptop-level training of models. It is accompanied by a rich example gallery of \href{https://pyshred-dev.github.io/pyshred/stable/examples/index.html}{Jupyter Notebook and Google Colab tutorials}.  


\section{Mathematical Background} \label{sec:background}

The SHRED architecture leverages three well-established mathematical principles: (i) the separation of variables, which was first historically used by Leibniz to solve ODEs in the late 1600s and later popularized by Fourier for solving PDEs, (ii) a decoding only strategy which bypasses the ill-conditioned direct computation of an inverse (known since the 1960s~\cite{croz1992stability}) required of encoder-decoder pair, and (iii) the 1981 Takens embedding theorem~\cite{takens,embedology} which shows that a time-delayed variable(s) is diffeomorphic to the original high-dimensional state space. 
In combination, these three critical concepts allow SHRED to robustly be used for sensing, model reduction and scientific discovery.  Moreover, the loss landscape of the SHRED models has been observed to be globally convex~\cite{gao2025sparse}.

A SHRED model is a neural network mapping from a trajectory of sensor measurements to a high-dimensional, spatio-temporal state. The architecture can be expressed as 
\begin{equation}
    \mathcal{H} \left( \{ \by_i \} _{i=t-k}^t\right) = \mathcal{F} \left( \mathcal{G} (\{ \by _i \}_{i=t-k}^t; \bW_{RN}); \bW_{SD} \right)
\end{equation}
where $\by_t$ consists of measurements of the high-dimensional state $\bx _t$, $\mathcal{F}$ is a fully-connected, feed-forward neural network parameterized by weights $\bW _{SD}$, and $\mathcal{G}$ is a sequence network model (e.g. LSTM, GRU, Transformer) parameterized by weights $\bW _{RN}.$  The desired network $\mathcal{H}^*$ minimizes the reconstruction loss, 
\begin{equation}
    \mathcal{H}^* \in \argmin_{\widetilde{\mathcal{H}} \in  \mathcal{H}} \sum _{i=1}^N ||\bx_i - \widetilde{\mathcal{H}}\left( \{ \by _j \}_{i-k}^i \right)||_2
\end{equation}
given a set of training states $\{ \bx_i \}_{i=1}^N$, corresponding measurements $\{ \by _i \} _{i=1}^N$. We train the network to minimize reconstruction loss.  A key innovation is the ability to train the SHRED model in a compressive representation~\cite{faraji2025shallow}. Thus, instead of training directly to the state-space $\bx _t$ - the data can first be compressed via the singular value decomposition~\cite{kutz2013data}, then trained to its compressed representation. This enables efficient laptop-level computing. In general, different strategies may be considered for data compression. See Appendix~\ref{sec:appendix_rom} for an example of compressive representation based on Fourier expansion.

\section{PySHRED Structure and Usage}
 \label{sec:code}

\begin{figure}[t]
   \centering
   \includegraphics[scale=0.4]{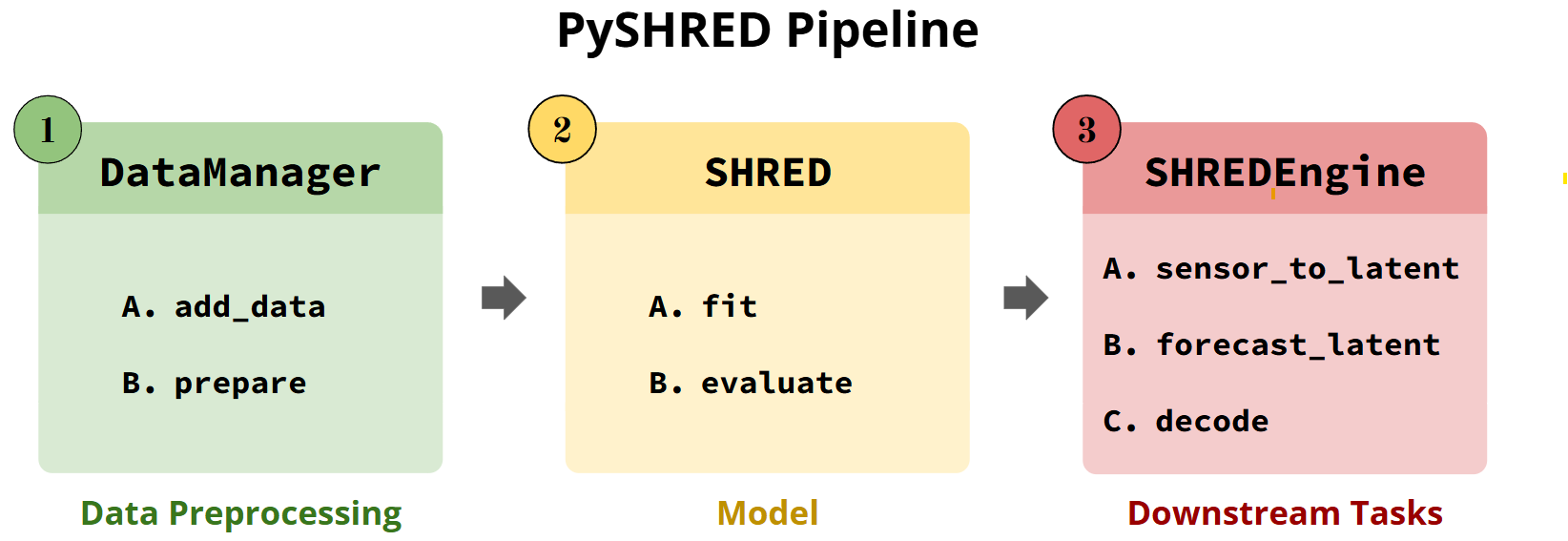}
   \caption{Overview of a basic PySHRED workflow, including the DataManager for preprocessing, the SHRED architecture for modeling, and the SHREDEngine for downstream tasks such as reconstruction and forecasting.}
   \label{fig:pyshred-pipeline}
\end{figure}

PySHRED is a modular and extensible package that streamlines the end-to-end SHRED pipeline, illustrated in Figure~\ref{fig:pyshred-pipeline}. Its design enables rapid experimentation and integration of custom components for a wide range of sensing and modeling tasks.  Specifically, SHRED provides a training framework that maps limited temporal (sequence) measurements to high-dimensional dynamical systems and/or spatiotemporal data from dynamical system snapshot observations.
The core of PySHRED comprises three primary modules (See Fig.~\ref{fig:pyshred-pipeline}):
\begin{itemize}
\item \texttt{DataManager} — Handles preprocessing and data partitioning into training, validation, and test sets. For parameterized systems, \texttt{ParametricDataManager} is available.
\item \texttt{SHRED} — Implements the SHRED architecture, with interchangeable components for the sequence model, decoder model, and optional latent forecaster, as shown in Figure~\ref{fig:mix-and-match}.
\item \texttt{SHREDEngine} — Provides a simple interface for performing downstream tasks including full-state reconstruction, long-horizon forecasting, and performance evaluation. For parametric systems, \texttt{ParametricSHREDEngine} is available.
\end{itemize}
Its modular design and simple interface enable researchers across domains to quickly build and experiment with a wide range of SHRED architectures.


\subsection{DataManager Module}
There are two types of modules responsible for data preprocessing and generating train/validation/test datasets. The \texttt{DataManager} is for non-parametric systems and the \texttt{ParametricDataManager} module is for parametric systems.
Both modules share the same initialization parameters and methods. Initialization involves specifying the train/validation/test split proportions, as well as the lags, which represent the number of timesteps included in each sensor sequence passed as input to \texttt{SHRED}.  These modules provide two main methods: \texttt{add\_data} and \texttt{prepare}. The \texttt{add\_data} method takes the arguments \texttt{data}, \texttt{id}, \texttt{compress}, along with sensor-related arguments.

\begin{itemize}
    \item \texttt{data} argument:
    \begin{itemize}
        \item \texttt{DataManager}: the \texttt{data} argument is an array-like object with time on the first axis and space on the remaining axes.
        \item  \texttt{ParametricDataManager}: the \texttt{data} argument is an array-like object with trajectories/experiments on first axis, time on the second axis, and space on the remaining axes.
    \end{itemize}
    \item \texttt{id} argument specifies a string to uniquely identify the dataset.
    \item \texttt{compress} argument specifies the number of modes saved from randomized SVD.
\end{itemize}
To add additional dynamically coupled fields (for multi-field SHRED), repeatedly call \texttt{add\_data} with additional datasets.  Once all data has been added, call \texttt{prepare} to obtain the train, validation, and test datasets used for training \texttt{SHRED}.

\subsection{SHRED Module}
\begin{figure}[t]
   \centering
   \includegraphics[scale=0.8]{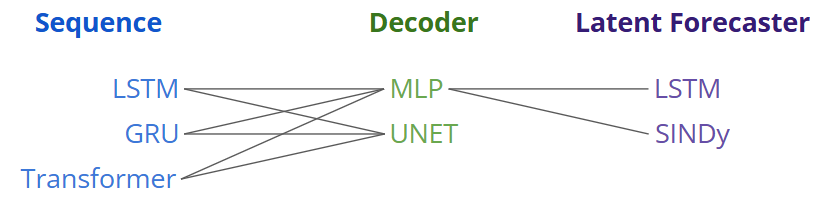}
   \caption{Available combinations across different models and architectures. As of PySHRED v1.0 the latent forecaster model is only available for the non-parametric regime.}
   \label{fig:mix-and-match}
\end{figure}

The \texttt{SHRED} module represents the core SHRED neural network architecture. It is composed of three models: sequence, decoder, and latent forecaster.
\begin{itemize}
    \item The sequence model encodes the temporal dynamics of sensor measurement trajectories into the latent space.
    \item The decoder model maps the latent space back to the high-dimensional full-state space.
    \item The latent forecaster model predicts future latent states in the absence of new sensor measurements.
\end{itemize}
Each model can be selected from a list of preset architectures by providing a string name, or customized by initializing model modules with user-defined settings. Different architectures for each model can be mixed and matched, allowing for flexible combinations tailored to specific applications. The architecture combinations available in PySHRED v1.0 are summarized in Figure~\ref{fig:mix-and-match}. Once \texttt{SHRED} is initialized, call the \texttt{fit} method with the train and validation datasets to train the model. Model performance can be assessed using the \texttt{evaluate} method, which computes mean squared error (MSE) on train, validation, or test datasets.

\subsection{SHREDEngine Module}
There are two types of modules responsible for downstream tasks including full-state reconstruction and forecasting. The \texttt{SHREDEngine} is for non-parametric systems and the \texttt{ParametricSHREDEngine} module is for parametric systems. Initialization involves passing in a \texttt{DataManager} or \texttt{ParametricDataManager} along with a trained \texttt{SHRED} model.  These modules provide four main methods:
\begin{itemize}
    \item \texttt{sensor\_to\_latent}: Generates the latent space from raw sensor measurements (Figure~\ref{fig:encoding}).
    \item \texttt{forecast\_latent}: Predicts future latent states starting from an initial sequence of latent vectors (Figure~\ref{fig:forecasting}). This method is only available for the \texttt{SHREDEngine} module.
    \item \texttt{decode}: Maps latent states back to the high-dimensional full-state space (Figure~\ref{fig:decoding}), and performs all post-processing steps, including unscaling, decompression, and multi-field separation.
    \item \texttt{evaluate}: Compares reconstructed full-state outputs against the ground truth, performing end-to-end evaluation in the physical space, including post-processing steps such as unscaling, decompression, and separating different fields. This \texttt{evaluate} method differs from that of \texttt{SHRED}, as it handles all post-processing steps (unscaling, decompression, multi-field separation) prior to evaluation.
\end{itemize}
The \texttt{DataManager}/\texttt{ParametricDataManager}, \texttt{SHRED}, and \texttt{SHREDEngine}/\texttt{ParametricSHREDEngine} modules work together to provide a simple and intuitive interface for running the SHRED pipeline.

\subsection{Example Usage}
Let \(X \in \mathbb{R}^{T \times m \times n}\) denote a spatiotemporal tensor with \(T\) snapshots, each an \(m \times n\) spatial field.  
The code chunk below walks through a minimal workflow: we (i) initialize a
\texttt{DataManager}, (ii) register \(X\) without compression and randomly select three sensor locations, (iii) split the data into training, validation, and test sets, (iv) train a default \texttt{SHRED} model, and
(v) initialize a \texttt{SHREDEngine} for downstream reconstruction and
forecasting tasks.
For full API details, see the \href{https://pyshred-dev.github.io/pyshred/stable/index.html}{documentation} and the \href{https://pyshred-dev.github.io/pyshred/stable/examples/index.html}{example gallery}.
\begin{lstlisting}[language=Python]
from pyshred import DataManager, SHRED, SHREDEngine

manager = DataManager()  # Initialize the data manager
manager.add_data(data=X, id="X", random=3, compress=False)  # Add data and randomly select 3 sensor locations
train, val, test = manager.prepare()  # Prepare datasets
shred = SHRED()  # Initialize the SHRED model
val_errors = shred.fit(train, val)  # Train SHRED
engine = SHREDEngine(manager, shred) # Initialize the SHRED engine
\end{lstlisting}

\section{Conclusion}
The PySHRED package in an open-source project that enables users with diverse backgrounds to apply SHRED in a user-friendly Pythonic environment. PySHRED is flexible in design, modular in structure, and extensible for new methods, enabling rapid experimentation and integration of custom models. We hope that through this work and through future works like this, PySHRED can continue to serve as a practical tool for sparse sensing, model reduction, and scientific discovery.

\section*{Acknowledgements}
We wish to acknowledge the support of the National Science Foundation AI Institute in Dynamic Systems grant 2112085. 
This work was additionally supported by the AFOSR/AFRL Center of Excellence in Assimilation of Flow Features in Compressible Reacting Flows under award number FA9550-25-1-0011, monitored by Dr. Chiping Li and Dr. Ramakanth Munipalli.

\newpage
\bibliographystyle{unsrtsiam}
\bibliography{refs}

\newpage

\appendix
\section*{Appendix}
\section{Sensing}
This example provides an easy on-ramp to using the PySHRED package by illustrating how SHRED can be used in sensing applications. The considered dataset is weekly mean sea-surface temperature and can be found at https://psl.noaa.gov/data/gridded/data.noaa.oisst.v2.html. We being by loading the data and applying a binary mask to consider only spatial locations corresponding to sea surface.
\begin{lstlisting}[language=Python, caption=Load sea-surface temperature data.]
import xarray as xr # for loading NetCDF files
import numpy as np

# Load SST and land-sea mask datasets
sst_dataset = xr.open_dataset('sst.wkmean.1990-present.nc')
mask_dataset = xr.open_dataset("lsmask.nc")

# Extract raw SST values and land-sea mask
data = sst_dataset["sst"].values
mask = np.squeeze(mask_dataset["mask"].values)

# Apply land-sea mask: set land regions (mask == 0) to 0
data[:,mask == 0] = 0

\end{lstlisting}

The next step is instantiating a \textit{DataManager} object from the PySHRED package. This manager allows us to easily perform a train/validation/test split and define the length of the temporal sequence used by the SHRED model during training. 
\begin{lstlisting}[language=Python, caption=Instantiate data manager.]
from pyshred import DataManager, SHRED, SHREDEngine
manager = DataManager(
    lags=52,          # 1 year of weekly history as input
    train_size=0.8,   # 80% for training
    val_size=0.1,     # 10% for validation
    test_size=0.1     # 10% for testing
)    
\end{lstlisting}

SHRED can be used with both stationary and mobile sensors. Here, we define trajectories for 3 sensors. The first sensor is mobile and charts a circular course around the middle of the Pacific ocean. The second two sensors are stationary, remaining fixed at one spatial coordinate in the Indian and Atlantic Oceans, respectively. The DataManager expects sensor trajectories to be defined as a list containing lists of tuples, hence the additional preprocessing.
\begin{lstlisting}[language=Python, caption=Define sensor locations.]
traj1 = np.array([90,200]) + 10*np.sin(np.array([np.arange(1727)*0.5, np.arange(1727)*0.5 + np.pi/2])).T
traj1 = traj1.astype(int)
traj2 = np.array([(100,100)] * 1727)
traj3 = np.array([(60,330)] * 1727)
traj1 = [(traj1[i,0], traj1[i,1]) for i in range(len(traj1))]
traj2 = [(traj2[i,0], traj2[i,1]) for i in range(len(traj2))]
traj3 = [(traj3[i,0], traj3[i,1]) for i in range(len(traj3))]
mobile = [traj1, traj2, traj3]
manager.add_data(
    data=data, 
    id="SST", 
    mobile=mobile, 
    compress=False 
)
    
\end{lstlisting}

Now that we've defined our training and testing splits as well as sensor locations, we can prepare our datasets and train the SHRED model. In this case, only 20 or so epochs are needed to achieve reasonable performance.

\begin{lstlisting}[language=Python, caption=Train SHRED model.]
train_dataset, val_dataset, test_dataset= manager.prepare()
shred = SHRED(
    sequence_model="LSTM",
    decoder_model="MLP",
    latent_forecaster="LSTM_Forecaster"
)
val_errors = shred.fit(
    train_dataset=train_dataset,
    val_dataset=val_dataset,
    num_epochs=20,
    verbose=False
)

\end{lstlisting}

\begin{figure}[t]
    \centering
    \includegraphics[width=0.9\linewidth]{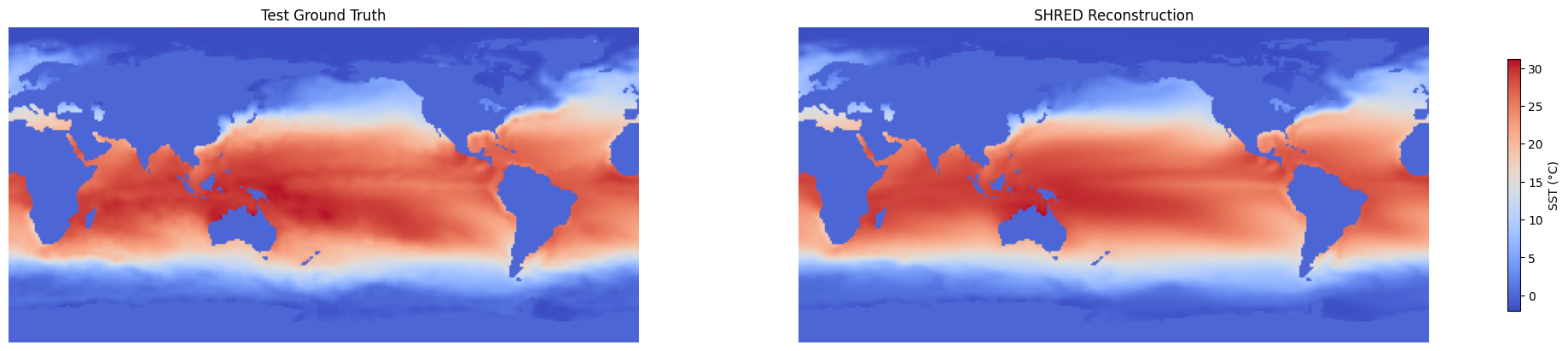}
    \caption{SHRED reconstruction of SST data.}
    \label{fig:shredsensingex}
\end{figure}

Once the SHRED model is trained, we instantiate a \textit{SHREDEngine} to evaluate the reconstruction performance in the test set. We first take the test set sensor measurements to generate the corresponding latent state sequence. From there, we apply the \textit{decode} method to map to the high-dimensional field. An example reconstruction from running the code in this example is shown in Fig. \ref{fig:shredsensingex}.

\begin{lstlisting}[language=Python, caption=Generate reconstructions.]
engine = SHREDEngine(manager, shred)
# obtain latent space of test sensor measurements
test_latent_from_sensors = engine.sensor_to_latent(manager.test_sensor_measurements)
test_reconstruction = engine.decode(test_latent_from_sensors)
\end{lstlisting}

\section{Reduced-ordered modeling}
\label{sec:appendix_rom}

This subsection explains in detail how to build parametric reduced order models based on shallow recurrent decoder networks, namely SHRED-ROM \cite{tomasetto2025shredrom}, within the PySHRED package. Specifically, we aim at modeling and reconstructing the double gyre flow \cite{SHADDEN2005271}, a time-dependent model for the interaction of two counter-rotating vortices that can lead to chaotic particle trajectories, in multiple parametric scenarios. For the sake of simplicity, we consider a rectangular domain $[0, L_x] \times [0, L_y]$, with $L_x = 2$ and $L_y = 1$, and the analytical velocity field $\mathbf{v} = [u, v]^T$ given by
\begin{align*}
u(x, y, t) &= -\pi I \sin\left( \pi f(x, t) \right) \cos\left( \pi y \right)
\\
v(x, y, t) &= \pi I \cos\left( \pi f(x, t) \right) \sin\left( \pi y \right) \frac{\partial f}{\partial x}
\end{align*}
where $I=0.1$ is the intensity parameter, $f(x, t) = \epsilon \sin(\omega t) x^2 + (1 - 2\epsilon \sin(\omega t)) x $, $\epsilon$ and $\omega$ are the perturbation amplitude and the frequency of the oscillation, respectively.

To handle the variability of perturbation amplitude and the oscillation frequency in a parametric ROM fashion, we build a collection of spatio-temporal snapshots by sampling the velocity field on a $50 \times 25$ spatial grid, while considering a uniform discretization of the time interval $[0,10]$ with step $\Delta t = 0.05$, for $100$ different parameter combinations (randomly picked). We then {\em (i)} pre-process the parametric trajectories by extracting smaller time-series with lag equal to $25$, {\em (ii)} we randomly select $3$ sensors measuring the horizontal velocity only, {\em (iii)} we split the data into training-validation-test sets, and {\em (iv)} we project the horizontal and vertical velocity through Proper Orthogonal Decomposition (POD) computed via singular values decomposition, by taking into account $4$ modes for each scalar field.

\begin{lstlisting}[language=Python, caption=Manage parametric data with POD-based compression.]
manager_pod = ParametricDataManager(
    lags = 25,
    train_size = 0.8,
    val_size = 0.1,
    test_size = 0.1,
    )

manager_pod.add_data(
    data=U,        # 3D array (parametric_trajectories, timesteps, field_dim)
    id="U",        # Unique identifier for the dataset
    random=3,      # Randomly select 3 sensor locations
    compress=4     # Spatial compression
)

manager_pod.add_data(
    data=V,         
    id="V",        
    compress=4     
)

manager_pod.add_data(
    data=parameters,
    id='mu',
    compress=False,
)

train_dataset, val_dataset, test_dataset = manager_pod.prepare()

\end{lstlisting}

Notice that the number of POD modes to retain may be selected by looking at the singular values decay or the POD reconstruction error. Moreover, the parameters may also be added to the DataManager, in order to provide additional inputs or to provide parameter estimations. 

Real-world sensor measurements are usually corrupted by noise. In this synthetic test case, we artificially add Gaussian random noise to sensor data.

\begin{lstlisting}[language=Python, caption=Synthetic noisy data.]

noise_std = 0.005
random_noise = np.random.normal(loc=0, scale=noise_std, size=manager_pod.sensor_measurements_df.shape)

manager_pod.sensor_measurements_df += random_noise

\end{lstlisting}

It is now possible to define and train SHRED-ROM in order to reconstruct spatio-temporal data for test parametric scenarios.

\begin{lstlisting}[language=Python, caption=SHRED-ROM training with POD-based compression.]

shred_pod = SHRED(
    sequence_model="LSTM",
    decoder_model="MLP",
    latent_forecaster=None
)

val_errors_shredpod = shred_pod.fit(
    train_dataset=train_dataset,
    val_dataset=val_dataset,
    num_epochs=100,
    patience=50,
    verbose=True,
)

\end{lstlisting}

Different compressive training strategies may be considered in place of POD, which provides a low-dimensional data-driven basis capable of explaining the variability of high-dimensional data. For example, physics-based or data-uninformed basis may be selected, such as spherical harmonics and trigonometric functions. After projecting the snapshots onto Fourier modes, we reduce the data dimensionality in the frequency domain by retaining only the terms corresponding to the lowest frequencies. Compared to POD compression, the resulting projected state dimension may be bigger, but the basis used for projection is general and not computed from data.

\begin{lstlisting}[language=Python, caption=Manage parametric data with Fourier-based compression.]
manager_fourier = ParametricDataManager(
    lags = 25,
    train_size = 0.8,
    val_size = 0.1,
    test_size = 0.1
)

manager_fourier.add_data(
    data=np.real(U_proj),
    id="Ufft_real",        
    random=3,
    compress=False
)

manager_fourier.add_data(
    data=np.imag(U_proj),        
    id="Ufft_imag",        
    compress=False
)

manager_fourier.add_data(
    data=np.real(V_proj),        
    id="Vfft_real",        
    compress=False
)

manager_fourier.add_data(
    data=np.imag(V_proj),        
    id="Vfft_imag",        
    compress=False
)

manager_fourier.add_data(
    data=parameters,
    id='mu',
    compress=False,
)

train_dataset,val_dataset,test_dataset = manager_fourier.prepare()

\end{lstlisting}

Notice that, if the sensor measurements of the high-dimensional field are already available, it is possible to directly provide them as additional input when adding the data to the DataManager.

\begin{lstlisting}[language=Python, caption=Manage parametric data with sensor measurments in input.]
manager_fourier.add_data(
    data=np.real(U_proj),
    id="Ufft_real",        
    measurements=measurements,
    compress=False
)
\end{lstlisting}

Similarly to the POD-based compressive training strategy, we can now define and train SHRED-ROM.

\begin{lstlisting}[language=Python, caption=SHRED-ROM training with Fourier-based compression.]

shred_fourier = SHRED(
    sequence_model="LSTM",
    decoder_model="MLP",
    latent_forecaster=None
)

val_errors_shredfourier = shred_fourier.fit(
    train_dataset=train_dataset,
    val_dataset=val_dataset,
    num_epochs=100,
    patience=50,
    verbose=True,
)

\end{lstlisting}

\begin{figure}[t]
    \centering
    \begin{overpic}[width=1\linewidth]{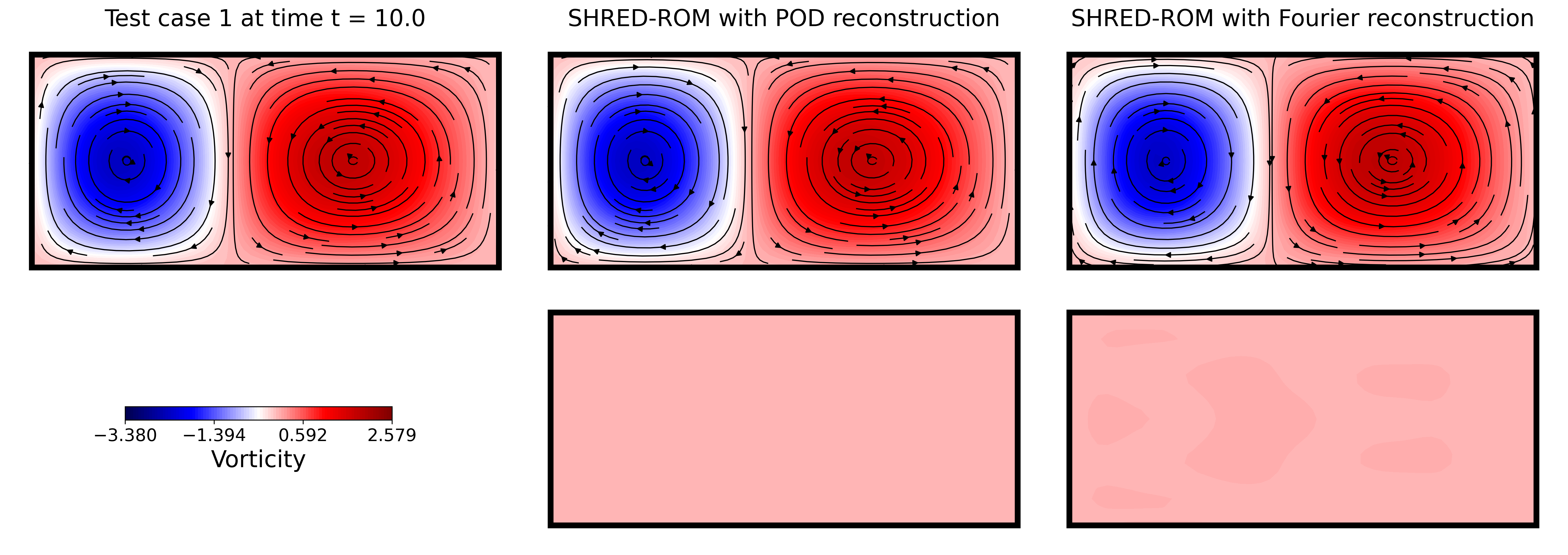}
    \end{overpic}
    \begin{overpic}[width=1\linewidth]{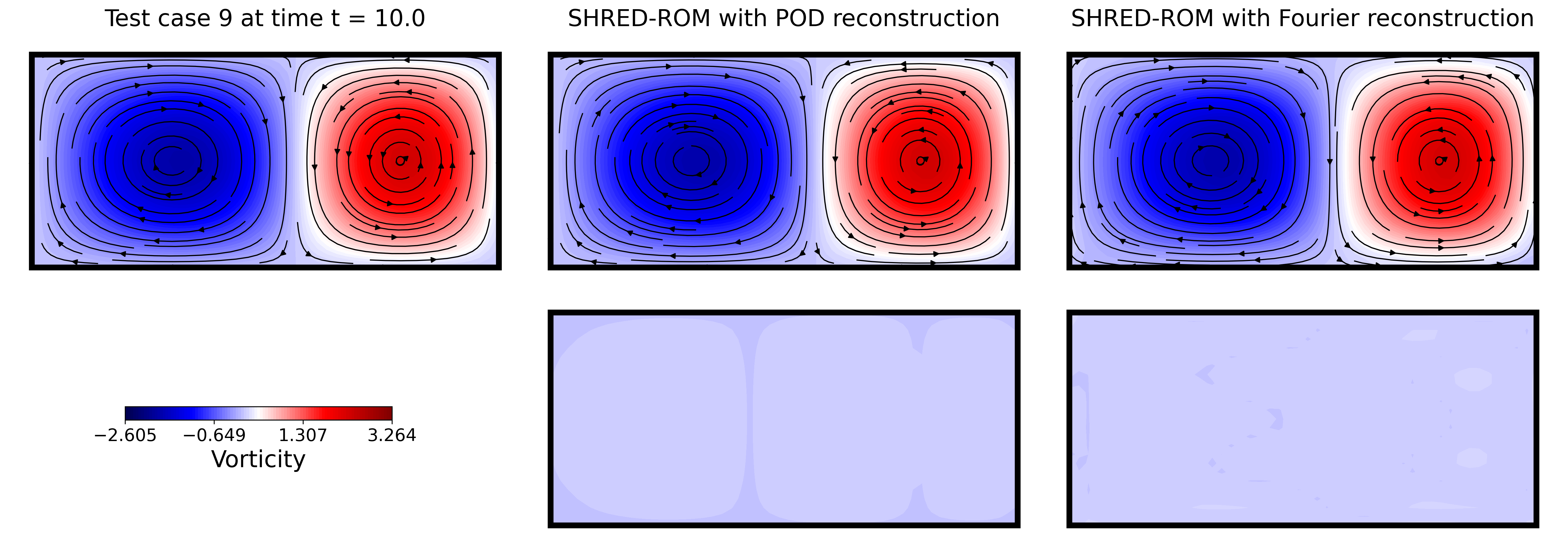}
    \end{overpic}
    \caption{Test vorticity fields and velocity contour plots (first column), corresponding SHRED-ROM reconstructions and residual fields when considering POD-based (second column) and Fourier-based (third column) compressive training.}
    \label{fig: doublegyre-contours}
\end{figure}

\begin{table}[h]
\centering
\renewcommand{\arraystretch}{1.3}
\normalsize

\begin{tabular}{|c|c|c|}
\hline
  & $u$ & $v$ \\
\hline
POD & 0.0347 & 0.0379 \\
\hline
Fourier & 0.0403 & 0.0609 \\
\hline
\end{tabular}
\caption{Mean relative error on test data committed by SHRED-ROM when reconstructing the horizontal and vertical velocity, i.e. $u$ and $v$, with POD-based and Fourier-based compressive training.}
\label{tab: doublegyre-errs}
\end{table}

After training, it is possible to assess the SHRED-ROM accuracy on parametric settings unseen during training. Figure \ref{fig: doublegyre-contours} shows the velocity contour plots and the vorticity field $w=-\frac{\partial u}{\partial y}+\frac{\partial v}{\partial x}$ of two test snapshots (first column), along with the corresponding SHRED-ROM reconstructions with POD-based (second column) and Fourier-based (third column) compressive training. Specifically, it is possible to assess the high accuracy of SHRED-ROM in both settings, with residual fields close to $0$, even though the former compression is slightly more accurate than the Fourier one. Table \ref{tab: doublegyre-errs} reports, instead, the mean relative test error $\frac{1}{N_s^{test}} \sum_{n=1}^{N_s^{test}}\frac{\|\boldsymbol{\psi}_n - \hat{\boldsymbol{\psi}}_n\|_2}{\|\boldsymbol{\psi}_n\|_2} $, where the vector $\boldsymbol{\psi}_n$ collects the spatial values of the $n$-th test snapshot of the generic field $\psi$ (either $u$ or $v$ in this context), while $\hat{\boldsymbol{\psi}}_n$ stands for the corresponding SHRED reconstruction. 

\section{Physics discovery via SINDy-SHRED}

In this subsection, we discuss physics discovery using SINDy-SHRED~\cite{gao2025sparse}.  The PySHRED package includes SINDy-SHRED to perform data-driven scientific discovery. 
We follow the same usage of DataManager to load the SST data and randomly select $50$ random sensors with lag equal to $52$. 

\begin{lstlisting}[language=Python, caption=Load SST data via DataManager. ]
# Loading data into DataManager with train, val, test split
manager = DataManager(
    lags = 52, train_size = 0.8, val_size = 0.1, test_size = 0.1,
)
# Adding the global SST data with 50 random sensors
manager.add_data(
    data = sst_data, id = "SST", random = 50, compress=False,
)
\end{lstlisting}

Then, we initialize a latent SINDy forecaster. This latent space forecaster predicts the latent space via the learned ODE in the SINDy unit. One could obtain the latent space prediction via forward ODE integration. We setup SHRED with the SINDy latent forecaster to (i) enable SINDy loss during training; (ii) enable physics-based prediction as in SINDy-SHRED; and (iii) perform scientific discovery from data~\cite{gao2025sparse}.  
\begin{lstlisting}[language=Python, caption=Setup the latent space forecaster using SINDy. ]
# Specify latent forecaster to be SINDy with linear terms
latent_forecaster = SINDy_Forecaster(poly_order=1, 
                    include_sine=True, dt=1/5)
# Train SHRED as regular with SINDy latent forecaster
shred = SHRED(sequence_model="GRU", decoder_model="MLP", 
                    latent_forecaster=latent_forecaster)
\end{lstlisting}

In SINDy-SHRED training, we apply a thresholding procedure with default threshold for every 20 \text{sindy\_thres\_epoch} epochs. 
Specifically, the training algorithm masks to zero all coefficients with absolute values below $0.05$ in every 20 epochs. 
This strategy is commonly used in sparse deep learning inference in gradient descent settings which is equivalent to $\ell_0$ regression with $\ell_2$ regularization~\cite{zheng2014high,gao2023convergence}. 
The SINDy regularization strength controls the level of constraint on how much we enforce the latent space of GRU into a consistent differential equation. 

\begin{lstlisting}[language=Python, caption=Training SINDy-SHRED using the regular fit function. ]
# Train SINDy-SHRED with regularization strength 1 and 
# perform thresholding for every 20 training epochs
val_errors = shred.fit(train_dataset=train_dataset, val_dataset=val_dataset, num_epochs=100, sindy_thres_epoch=20, sindy_regularization=1)
\end{lstlisting}

Based on the selection of random sensors, we discover is a linear model for the global sea surface temperature data that 
\begin{align*}
\dot{x}_0 &= 0.048 - 0.122\,x_0 - 0.279\,x_1 - 0.103\,x_2,\\
\dot{x}_1 &= 0.012 + 0.066\,x_0 + 0.036\,x_1 + 0.070\,x_2,\\
\dot{x}_2 &= -0.165 - 0.002\,x_0 + 0.218\,x_1 - 0.159\,x_2.
\end{align*}

\begin{figure}[t]
    \centering
    \includegraphics[width=0.9\linewidth]{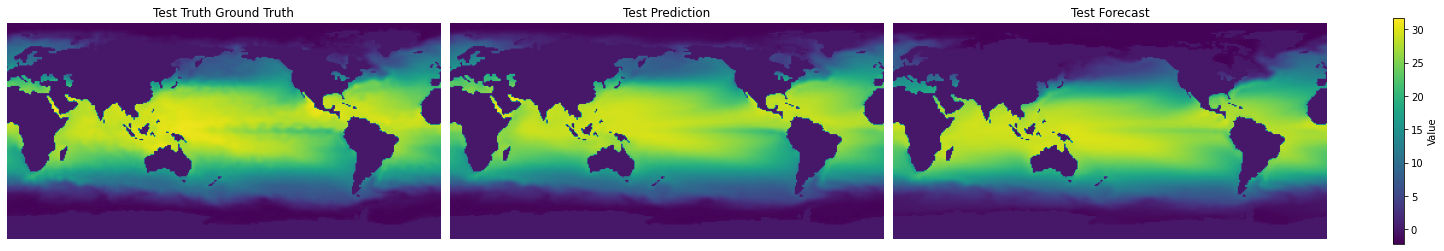}
    \caption{SINDy-SHRED reconstruction and forecast of SST dynamics.}
    \label{fig:sst_sindy_shred}
\end{figure}

By unrolling the ODE forward, we obtain the future latent states autoregressively, and we predict the spatial information from the predicted latent space. 


\section{SHREDEngine Pipelines}

\begin{figure}
   \centering
   \includegraphics[scale=0.3]{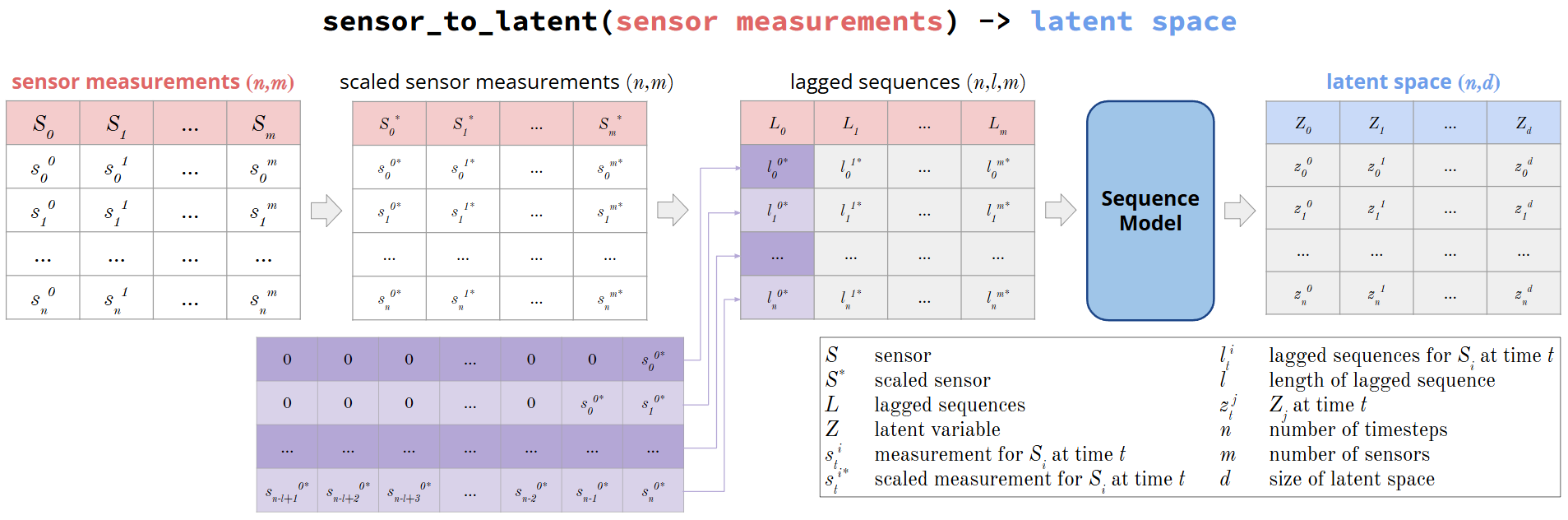}
   \caption{Sensor to latent pipeline: sensor measurements are transformed into latent representations via the \texttt{sensor\_to\_latent} method.}
   \label{fig:encoding}
\end{figure}

\begin{figure}[h]
   \centering
   \includegraphics[scale=0.45]{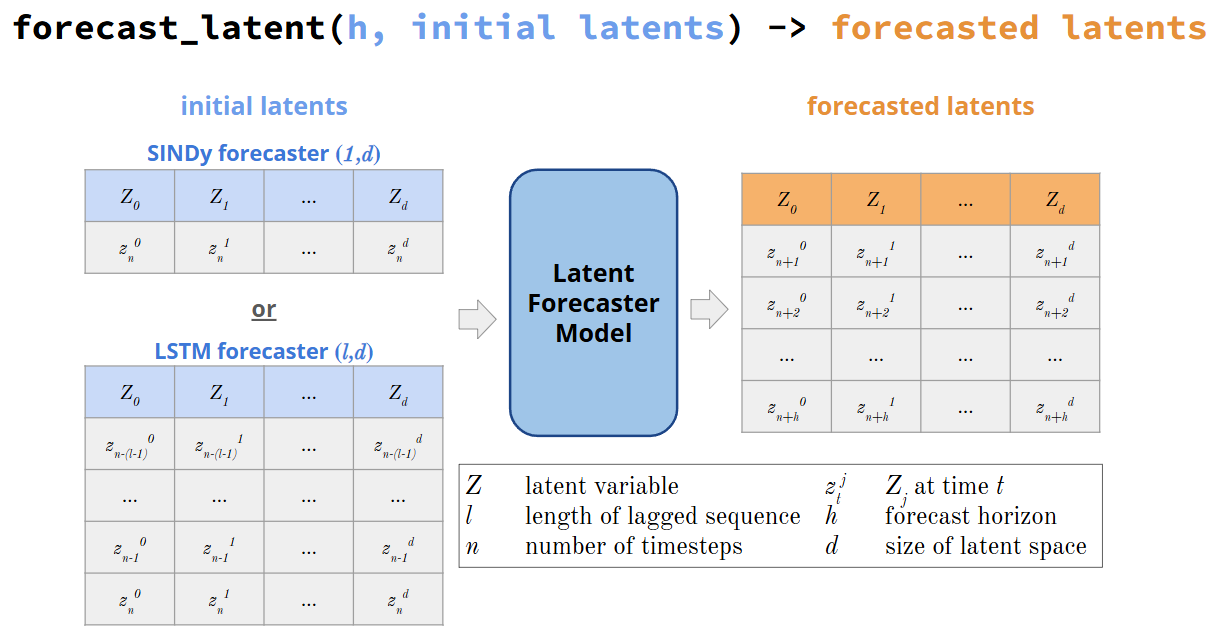}
   \caption{Forecasting pipeline: a seed sequence of latent vectors is propagated forward in time using the \texttt{forecast\_latent} method to obtain future latent states, which can subsequently be decoded to full‐state forecasts.}
   \label{fig:forecasting}
\end{figure}

\begin{figure}[h]
   \centering
   \includegraphics[scale=0.5]{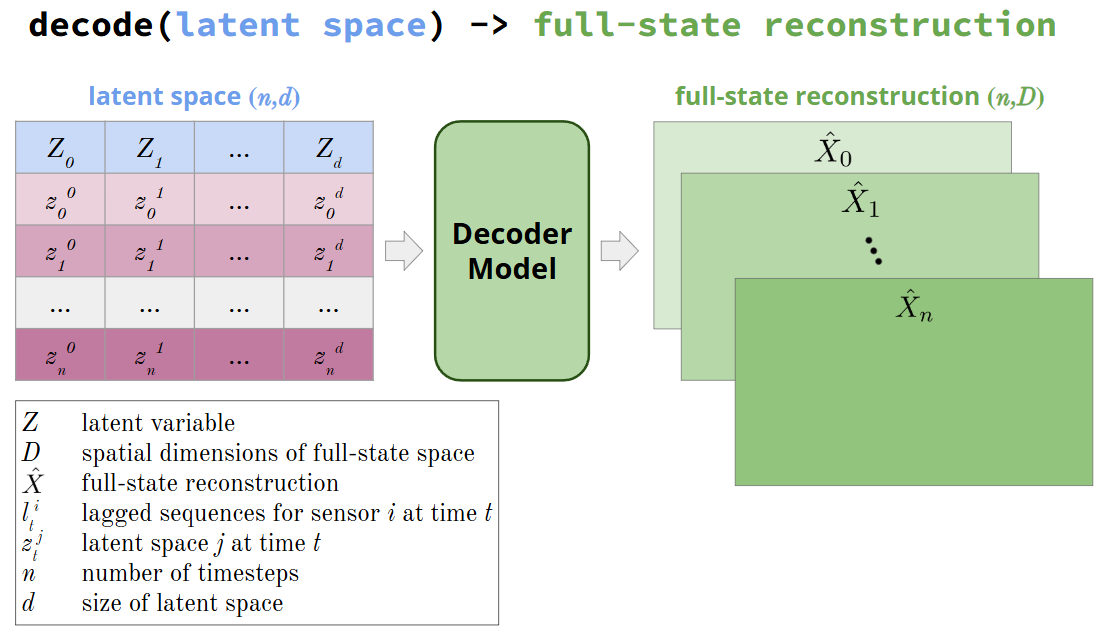}
   \caption{Decoding pipeline: latent trajectories are mapped back to the full
  high-dimensional state using the \texttt{decode} method. For brevity, the figure omits the additional post-processing performed by \texttt{decode}: unscaling, decompression, and multi-field separation.}
   \label{fig:decoding}
\end{figure}

Figures ~\ref{fig:encoding}, ~\ref{fig:forecasting}, and \ref{fig:decoding} illustrate the three core tasks handled by \texttt{SHREDEngine}: encoding sensors to latent space, forecasting latent trajectories, and decoding back to full-state
fields.

\end{document}